\documentclass[lettersize,journal]{IEEEtran}
\usepackage{amsmath,amsfonts}
\usepackage{algorithmic}
\usepackage{algorithm}
\usepackage{array}
\usepackage[caption=false,font=normalsize,labelfont=sf,textfont=sf]{subfig}
\usepackage{textcomp}
\usepackage{stfloats}
\usepackage{url}
\usepackage{verbatim}
\usepackage{graphicx}
\usepackage{cite}
\usepackage{xcolor}
\usepackage{multirow}
\usepackage{multicol}

\usepackage{numprint}
\npdecimalsign{.}
\nprounddigits{2}

\def\BibTeX{{\rm B\kern-.05em{\sc i\kern-.025em b}\kern-.08em
    T\kern-.1667em\lower.7ex\hbox{E}\kern-.125emX}}
\begin{document}

\title{WaveCatBoost for Probabilistic Forecasting of Regional Air Quality Data
\author{Jintu Borah,~\IEEEmembership{Graduate Student Member,~IEEE,}~Tanujit Chakraborty, Md. Shahrul Md. Nadzir, Mylene G. Cayetano, and Shubhankar Majumdar,~\IEEEmembership{Senior Member,~IEEE}}

\thanks{This work was supported in part by ASEAN-India Science and Technology Collaboration (AISTIC) Funding
from the Department of Science and Technology, India (CRD/2020/000320) for providing incentives.}
\thanks{J. Borah (email: jintub64@gmail.com) \& S. Majumdar (email: shubuit@gmail.com) is with the Dept. of ECE, NIT Meghalaya, India}
\thanks{T. Chakraborty (email: ctanujit@gmail.com) is with the Sorbonne University, Abu Dhabi, UAE and Sorbonne Center for Artificial Intelligence, Paris, France}
\thanks{M.S.M. Nadzir is with Dept. of Earth Sciences and Environment, Universiti Kebangsaan, Selangor, Malaysia. (e-mail: shahrulnadzir@ukm.edu.my).}
\thanks{M.G. Cayetano is with the Institute of Environmental Science and Meteorology, University of Philippines, Manila, Philippines (e-mail: mcayetano@iesm.upd.edu.ph).}
\thanks{J. Borah and T. Chakraborty (Joint first authors) have equal contributions.}
}



\maketitle
\begin{abstract}
Accurate and reliable air quality forecasting is essential for protecting public health, sustainable development, pollution control, and enhanced urban planning. This letter presents a novel WaveCatBoost architecture designed to forecast the real-time concentrations of air pollutants by combining the maximal overlapping discrete wavelet transform (MODWT) with the CatBoost model. This hybrid approach efficiently transforms time series into high-frequency and low-frequency components, thereby extracting signal from noise and improving prediction accuracy and robustness. Evaluation of two distinct regional datasets, from the Central Air Pollution Control Board (CPCB) sensor network and a low-cost air quality sensor system (LAQS), underscores the superior performance of our proposed methodology in real-time forecasting compared to the state-of-the-art statistical and deep learning architectures. Moreover, we employ a conformal prediction strategy to provide probabilistic bands with our forecasts. 

\end{abstract}
\section{Introduction}
Air pollution is a critical global concern, adversely affecting public health and the environment. The rapid expansion of industrialization and urbanization have surged the emission of air pollutants, resulting in significant challenges to sustainable living. In response to these concerns, the World Health Organization has issued guidelines for six major air pollutants and set National Ambient Air Quality Standards. These pollutants, including nitrogen dioxide ($NO_2$), ozone ($O_3$), carbon monoxide ($CO$), sulfur dioxide ($SO_2$), particulate matter with diameters of 2.5 mm or less ($PM_{2.5}$), and 10 mm or less ($PM_{10}$), originate from various sources such as transportation, industry, and natural processes. Despite considerable efforts to reduce emissions and lower ambient concentrations of these pollutants, air pollution remains a significant cause of mortality worldwide. \par
To mitigate these concerns related to polluted air, several countries have implemented real-time air quality forecasting systems \cite{shaban2016urban}. Accurate prediction of air pollutant concentration levels is paramount for public awareness campaigns, environmental management, and healthcare interventions, among many others. 
By leveraging these forecasts, government regulations and public policies can be designed to address pollution-related health concerns and promote sustainable development initiatives. In recent years, machine learning algorithms have emerged as powerful tools for enhancing the accuracy of air quality predictions \cite{du2019deep}. These computational techniques, driven by their ability to decipher complex patterns and relationships within vast datasets, offer a promising avenue for advancing pollutant forecasting models \cite{borah2023aicarebreath, gu2018recurrent}.\par However, existing methods encounter challenges adapting to diverse monitoring environments and accurately predicting pollutant concentrations in real-time \cite{JMLR:v23:21-1177}. Traditional models often struggle with the dynamic nature of air quality, hindered by limitations in handling non-linear relationships and non-stationary variations in pollutant dynamics. To address these challenges in the forecasting methodologies, wavelet-based transformation is widely applied in other applied domains, including epidemiology \cite{panja2023epicasting}, economics \cite{sengupta2023forecasting}, and others. These wavelet-based forecasting approaches transform non-stationary seasonal data into a series of uncorrelated distinct components. Subsequently, the component series are modeled with various statistical and deep learning \cite{sasal2022w} architectures to solve diverse problems in time series forecasting. While the accuracies of these wavelet-based deep learning architectures surpass traditional approaches for ultra-long time series, their training and prediction times obstruct the generation of real-time forecasts. To mitigate this challenge, decision tree-based boosting architectures are favored for their faster convergence time in the presence of large datasets \cite{grinsztajn2022tree}. Several boosting algorithms, including random forest, extreme gradient boosting, and light gradient boosting machines, offer higher accuracy compared to deep learning methods, but they overlook the sequential nature of the time series, resulting in target leakage \cite{liu2020spatial}. The recently developed CatBoost algorithm employs an ordered boosting methodology to overcome this issue \cite{prokhorenkova2018catboost}. \par Motivated by this background, in this letter, we develop a novel wavelet-based CatBoost (WaveCatBoost) model for generating accurate real-time forecasts of air pollutant concentration levels to mitigate the overarching challenges of reliable forecasting of air quality data. This integration aims to leverage the strengths of both techniques, providing a robust and versatile model capable of delivering reliable point and probabilistic forecasts across a range of air pollutants via deploying a conformal prediction approach. By conducting an extensive experimental evaluation and comparing existing models, we demonstrate the superiority of our proposed approach in advancing air quality forecasting and addressing the critical gaps present in the current literature.

\begin{figure}
    \centering
    \includegraphics[width=0.48\textwidth]{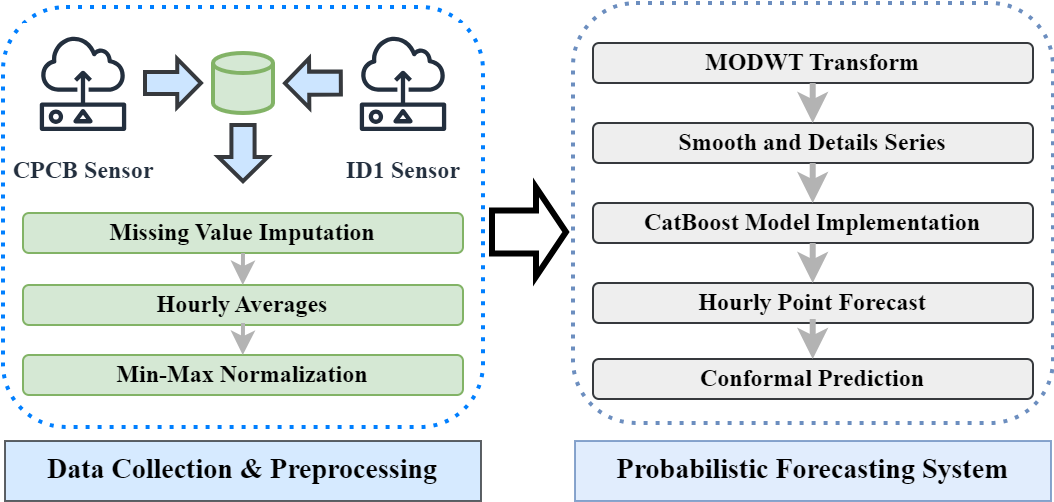}
    \caption{Pipeline of the proposed air quality forecasting system}
    \label{fig:flowchart}
\end{figure}
\begin{figure*}[h]
    \centering
    \includegraphics[width=\textwidth]{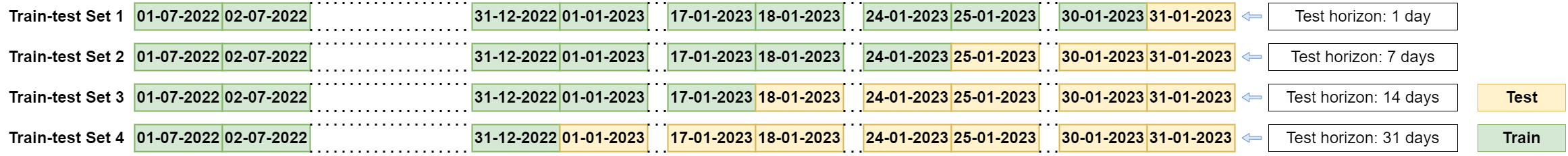}
    \caption{Depiction of the forecast horizons considered in this study}
    \label{fig:horizons}
\end{figure*}
\section{Material and Methodology}
\subsection{Data Collection and Preprocessing}
Real-time data on air pollutant concentration levels are gathered through two wireless sensors, the Central Air Pollution Control Board (CPCB) sensor network and a low-cost air quality sensor system named ID1, strategically positioned in Meghalaya, India. These air quality sensors continuously monitor the concentrations of NO$_2$ (in ppb), O$_3$ (in ppb), CO (in ppb), SO$_2$ (in ppb), PM$_{2.5}$ (in $\mu$g/m$3$), and PM$_{10}$ (in $\mu$g/m$3$) and transmit the data to the server at one-minute intervals. The transmitted information is sequentially stored in a tabular data format within the database for further analysis. We extract the raw data from the server database for a year, from July 1, 2022, to July 31, 2023. The preprocessing phase of our experiment involves adjusting the missing data entries and computing hourly averages to generate near-real-time data on the concentration of various air pollutants. Additionally, we normalize the dataset using the min-max normalization strategy, which is crucial for optimizing the convergence of data-driven forecasting algorithms used in our analysis. A visual representation outlining the processes of data collection, pre-processing, and analysis employed in this study is depicted in Fig. \ref{fig:flowchart}.

\subsection{Proposed WaveCatboost}
The key concept behind the wavelet-based CatBoost framework, which we named WaveCatBoost, involves employing the scale-invariant maximal overlapping discrete wavelet transformation (MODWT) \cite{percival2000wavelet} to denoise the air pollutant concentrations series and modeling them using an ensemble of CatBoost models. The MODWT decomposition is applied to filter the normalized air pollutant data $(Y_t)_{t=1}^N$ using a sequence of wavelet $\left\{p_m; m = 0, 1, \ldots, \mathcal{M}-1 \right\}$ and scaling $\left\{q_m; m = 0, 1, \ldots, \mathcal{M}-1 \right\}$ filters. These filters adhere to the even-length scaling assumption, $\sum_{m = 0}^{\mathcal{M}-1} p_m p_{m+2n} = \sum_{m = 0}^{M-1} q_m q_{m+2n} = 0$ for any non-zero integers $n$. Consequently, applying the pyramid algorithm \cite{percival2000wavelet} generates wavelet and scaling coefficients for the transformed time series, allowing the original series $Y_t$ to be represented as a set of equal-sized uncorrelated time series:
\begin{equation*}
    Y_t = \sum_{k= 1}^K D_{k, t} + S_{K, t} \quad t = 1, 2, \ldots, N,
\end{equation*}
where $K =  \log_e N$ is the number of decomposition levels applied, $D_{k, t} (k= 1, 2, \ldots, K)$ is the $k^{th}$ details series of $Y_t$, captures the local fluctuations of the series, and the smooth series $S_{K, t}$ preserves the low-frequency trend of $Y_t$. This iterative decomposition essentially helps filter the true signal from the noisy data and aids in modeling long-term dependent, non-stationary, and seasonal data by breaking it into several lower-resolution components. Subsequently, the task of generating $h$-step ahead forecasts of $Y_t$ based on its historical data can be solved by forecasting these low-resolution components based on their corresponding prior observation. In our proposed WaveCatBoost architecture, we generate the forecasts for each of the wavelet and scaling coefficients of $Y_t$ using the gradient boosting CatBoost \cite{prokhorenkova2018catboost} algorithm and ensemble them using an inverse MODWT (IMODWT) \cite{percival2000wavelet} approach as:
\begin{equation}
    \hat{Y}_{N+h} = \operatorname{IMODWT}\left(\hat{D}_{1,N+h}, \ldots, \hat{D}_{K,N+h}, \hat{S}_{K,N+h}\right),
    \label{Eq_IMODWT}
\end{equation}
$\text{with } \hat{D}_{k,N+h} = f\left(D_{k,1}, D_{k,2}, \ldots, D_{k,N}\right); k = 1, 2, \ldots, K$ and
$\hat{S}_{K,N+h}  = f\left(S_{K,1}, S_{K,2}, \ldots, S_{K,N}\right),$
where the function $f$ indicates the CatBoost model applied on each of the $p$ lagged values of details and smooth series. The CatBoost framework operates by iteratively combining weak symmetric decision trees into a strong regressor. Unlike traditional gradient boosting methods, it is prone to prediction shifts from target leakage. CatBoost architecture employs an ordered boosting mechanism. During the learning process, the training data undergoes a series of time-dependent permutations, denoted as $\left(\sigma_0, \sigma_1, \ldots, \sigma_u\right)$. The initial permutation $\sigma_0$ determines the leaf values, while the subsequent $u$ permutations decide the structure of the weak learners. At each iteration $v$, the algorithm considers a randomly chosen permutation $\sigma_r$. The supporting models $M_{r, j}$ are then fitted to the preceding $j$ observations of $\sigma_r$, generating the output $M_{r, j}(i)$ for the $i^{th}$ instance in $\sigma_r$. To construct the tree $T_v$ at the current iteration, gradients for the $i^{th}$ time-index are computed by considering previous time steps in the $\sigma_r$ permutation, i.e., $\operatorname{grad}_{r, \sigma(i)-1}(i)$ where 
\begin{equation}
    \operatorname{grad}_{r, j}(i) = \left. \frac{\partial L\left(y_i, A\right)}{\partial A}\right|_{A = M_{r, j}(i)}.
    \label{Eq_Grad}
\end{equation}
The leaf value for the $i^{th}$ instance, used for evaluating candidate splits, is obtained by averaging the corresponding gradients $\operatorname{grad}_{r, \sigma_r(i)-1}$ of preceding examples belonging to the same leaf node $\operatorname{Leaf}_{r\left(i\right)}$. Once the tree $T_v$ is constructed, it is used for boosting all the supporting models $M_{r',j}$ with varied sets of leaf values depending on the first $j$ instances in the $\sigma_{r'}$ permutation. When all the weak learners are constructed, the leaf value of the CatBoost model is calculated by the standard gradient boosting procedure. For generating the forecasts $\hat{D}_{k, N+h}; k = 1, 2, \ldots, K$ and $\hat{S}_{K, N+h}$, the leaf values are computed based on target statistics learned from the CatBoost architecture using lagged inputs. The algorithmic structure of the proposed WaveCatBoost model is given in Algorithm \ref{WaveBB_Algo}.
\subsection{Probabilistic Forecasting using Conformal Pediction}
Along with the point forecasts generated by the WaveCatBoost model, we focus on quantifying the uncertainty associated with these forecasts using the conformal prediction approach \cite{vovk2005algorithmic}. This non-parametric procedure generates a probabilistic band around the point forecasts based on a conformal score ($CS_t$). To calculate $CS_t$ corresponding to $Y_t$ \cite{angelopoulos2023conformal}, the architecture considers $p$ lagged values of the target series ($Y_{t-p}$) and applies the WaveCatBoost and an uncertainty model $\mathbf{\hat{U}}$ such that,
$
    CS_t = \frac{\left|Y_t - \operatorname{WaveCatBoost}\left(Y_{t-p}\right)\right|}{\mathbf{\hat{U}}\left(Y_{t-p}\right)}.
$
Following the sequential nature of $Y_t$ and the $CS_t$, the conformal quantile can be computed with a weighted aggregation method having a fixed window $w_t = \mathbb{1}\left(t' \geq t - \kappa\right), \forall \; t'< t$ of length $\kappa$ as
\begin{equation*}
    CQ_t = \operatorname{inf}\left\{q: \frac{1}{\operatorname{min\left(\kappa, t'- 1\right)}}\sum_{t' = 1}^{t-1} CS_{t'}w_t \geq 1 - \alpha \right\}.
\end{equation*}
The conformal prediction intervals based on these weighted conformal quantiles can be estimated as 
\begin{equation}
    \left[\operatorname{WaveCatBoost}\left(Y_{t-p}\right) \pm CQ_t \mathbf{\hat{U}}\left(Y_{t-p}\right)\right].
    \label{CP_WCB_Eq}
\end{equation}

\begin{algorithm}[h]
\caption{WaveCatBoost Model}
\begin{algorithmic}[1]
\REQUIRE Air pollutant concentration data $\left\{Y_t\right\}_{t=1}^N$
\ENSURE $h$-step ahead forecast $\left\{\hat{Y}_{N+1}, \ldots, \hat{Y}_{N+h}\right\}$
\STATE Initialize the MODWT algorithm with Haar filter and $K = \log_e N$ levels.
\STATE Transform the training data $\left\{Y_t\right\}$ using the MODWT approach into $K$ uncorrelated details series $\left\{D_{k,t}\right\}_{k=1}^{K}$ and a smooth series $\left\{S_{K,t}\right\}$
\FOR{$k  = 1, 2, \ldots, K$}
\STATE Apply CatBoost model to the $k^{th}$ detail series $D_{k,t}$.
\STATE $\sigma_0, \sigma_1, \ldots, \sigma_u \leftarrow $ Time-dependent permutations of $D_{k,t}$.
\STATE $\sigma_0$ decides the leaf nodes and the next $u$ permutations determine the structure of the symmetric weak learners.
\FOR{$v  = 1, 2, \ldots, V$}
\STATE $\sigma_r \longleftarrow \; \text{random permutations of} \; \left[\sigma_0, \sigma_1, \ldots, \sigma_u\right]$.
\STATE Fit $M_{r,j}$ to $j$ preceeding observations of $\sigma_r$ and generate output $M_{r,j}(i)$ for $i^{th}$ instance of $\sigma_r$.
\STATE Calculate leaf values of the week learner $T_v$ by averaging the gradients $\operatorname{grad}_{r, \sigma_r(i)-1}$ (Eq. \ref{Eq_Grad}) of previous examples belonging to the same node $\operatorname{Leaf}_{r\left(i\right)}$.
\STATE Boost the models $M_{r',j}$ of $\sigma_{r'}; \; r' \neq r$ using $T_v$.
\ENDFOR
\STATE Calculate leaf value for the CatBoost model by applying gradient boosting approach to $V$ week learners' leaf nodes and generate the one-step ahead forecast for $D_{k,t}$.
\STATE Iteratively generate multi-step ahead forecast.
\ENDFOR
\STATE Similarly, utilize a CatBoost architecture on $S_{K,t}$ to generate the corresponding one-step ahead and multi-step ahead forecast iteratively.
\STATE Apply Inverse MODWT transform on the component forecasts from the ensemble of CatBoost models and generate the final forecasts of the desired horizon as in Eq. \ref{Eq_IMODWT}.
\end{algorithmic}
\label{WaveBB_Algo}
\end{algorithm}

\section{Experimental Results}
In this section, we present experimental results showcasing the efficacy of our proposed WaveCatBoost framework compared to benchmark forecasting methods. We forecast hourly air pollution levels across different horizons through subsequent experiments. 
It helps identify each model's temporal sensitivity, guiding the selection of the most suitable model for specific prediction needs. Additionally, it ensures the models apply to a range of real-world scenarios, from immediate health advisories to long-term urban planning. This analysis informs resource allocation, enabling tailored responses for short-term and long-term forecasts, and supports the construction of ensemble models for enhanced accuracy. Furthermore, it deepens our scientific understanding of air quality dynamics and aids in crafting effective policies and mitigation strategies, addressing different timeframes, from rapid interventions to long-term sustainability planning. Our analysis utilizes a rolling window forecasting technique, assessing model performance across four distinct train-test scenarios: with forecast horizons set at 1 day (24 test points), 7 days, 14 days, and 31 days. Fig. \ref{fig:horizons} presents a graphical representation of these train-test configurations. 
In our analysis we compare the performance of the proposed WaveCatBoost architecture with state-of-the-art forecasting methodologies including NLinear, DLinear, light gradient-boosting machines (LGBM), extreme gradient-boosting (XGB), NBeats, NHiTS, Transformer, temporal convolution network (TCN), recurrent neural network (RNN), gated recurrent unit (GRU), long-short-term memory (LSTM), temporal fusion transformer (TFT), and CatBoost model \cite{JMLR:v23:21-1177}. We measure the forecast accuracy of various frameworks using the mean absolute scaled error (MASE) metric, one of the most recommended performance measures in forecasting literature, on the test data as follows:
\begin{equation*}
\text{MASE} = \frac{N-S}{h} \frac{\sum_{t=N+1}^{N+h}\left| \hat{Y}_t - Y_t \right|}{\sum_{t=S+1}^{N} \left|Y_{t} - Y_{t-S} \right|},
\end{equation*}
where $h$ is the forecast horizon, $N$ is the length of training data, $\hat{Y}_t$ is the forecast of $Y_t$ at time $t$ and $S$ is the seasonality of the data.

\begin{figure}
    \centering
    \subfloat[]{\includegraphics[width=0.23\textwidth]{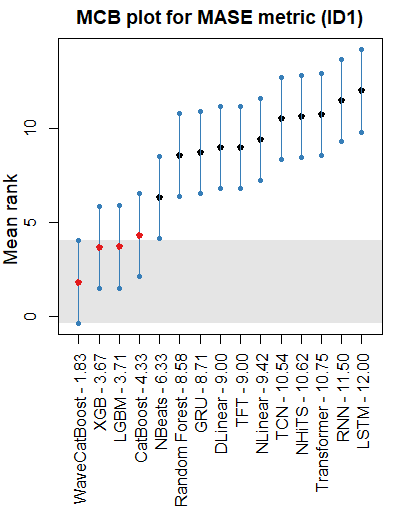}}
    \label{fig:mcb_id1}
    \hfil
    \subfloat[]{\includegraphics[width=0.23\textwidth]{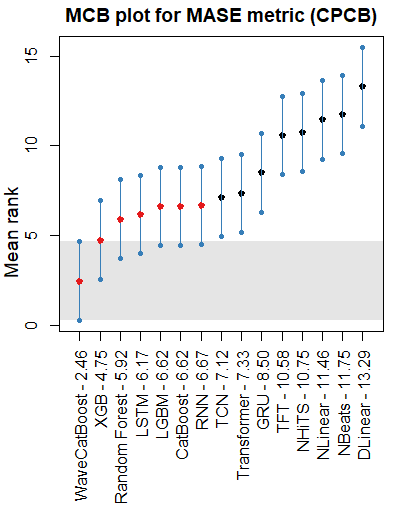}}
    \label{fig:mcb_cpcb}
    \caption{MCB plots for MASE metric of the various forecasting models across all forecast horizons utilizing two different sensors (a) ID1, (b) CPCB}
    \label{fig:mcb fh}
\end{figure}
\begin{figure}
    \centering
    \subfloat[]{\includegraphics[width=.23\textwidth]{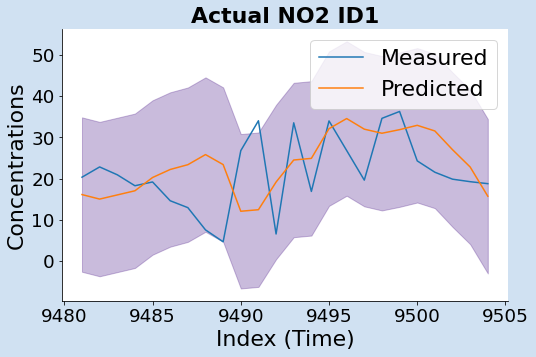}}
    \label{fig:cp_id1_no2_1d}
    \hfil
    \subfloat[]{\includegraphics[width=.23\textwidth]{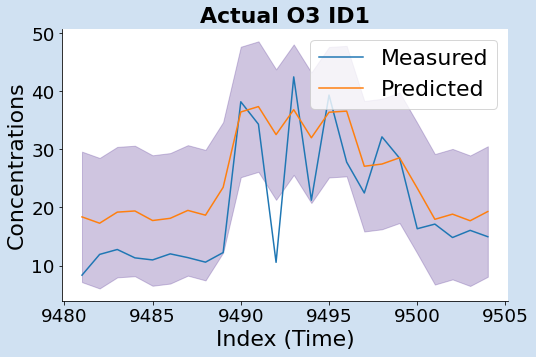}}
    \label{fig:cp_id1_o3_1d}
    \hfil
    \subfloat[]{\includegraphics[width=.23\textwidth]{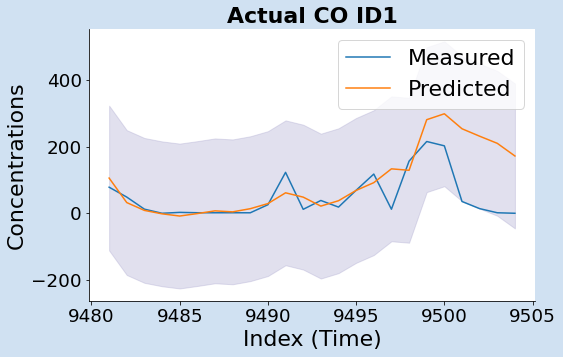}}
    \label{fig:cp_id1_co_1d}
    \hfil
    \subfloat[]{\includegraphics[width=.23\textwidth]{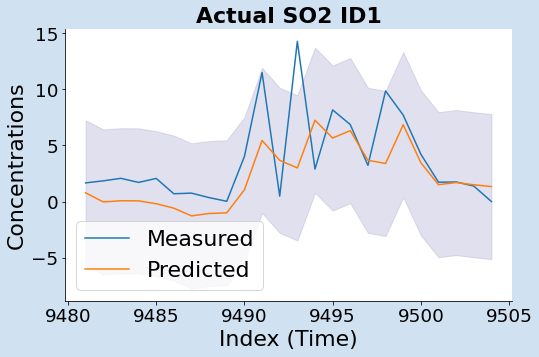}}
    \label{fig:cp_id1_so2_1d}
    \hfil
    \subfloat[]{\includegraphics[width=.23\textwidth]{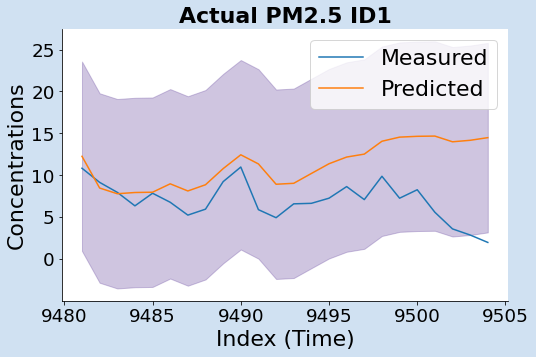}}
    \label{fig:cp_id1_pm2_1d}
    \hfil
    \subfloat[]{\includegraphics[width=.23\textwidth]{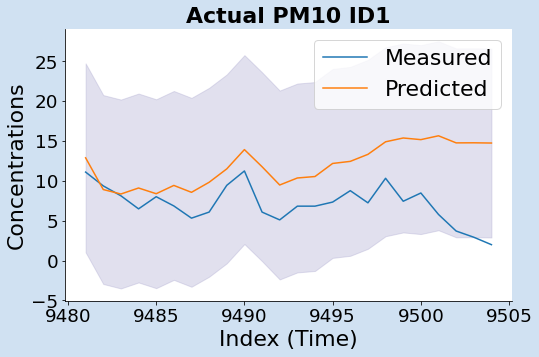}}
    \label{fig:cp_id1_pm10_1d}
    \caption{Actual vs Forecasts of the targeted pollutants for ID1 Sensor for 1 day forecast horizon (a) $NO_2$, (b) $O_3$, (c) $CO$, (d) $SO_2$, (e) $PM_{2.5}$, and (f) $PM_{10}$}
    \label{fig:cp 1d id1}
\end{figure}
\begin{figure}
     \centering
     \subfloat[]{\includegraphics[width=.23\textwidth]{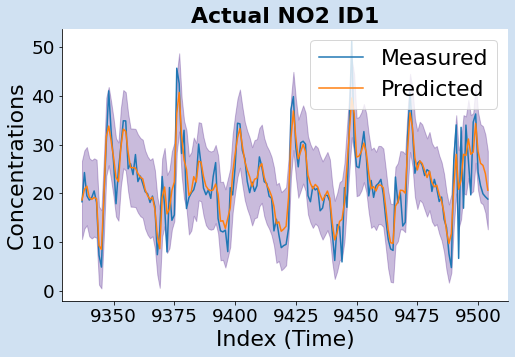}}
    \label{fig:cp_id1_no2_7d}
    \hfil
    \subfloat[]{\includegraphics[width=.23\textwidth]{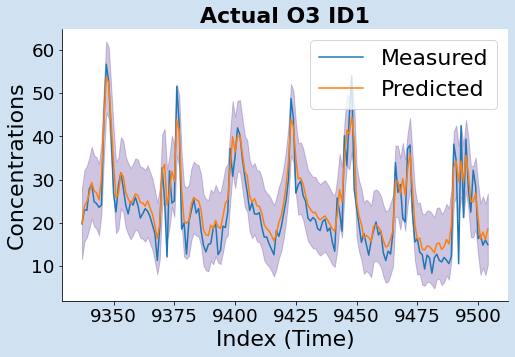}}
    \label{fig:cp_id1_o3_7d}
    \hfil
    \subfloat[]{\includegraphics[width=.23\textwidth]{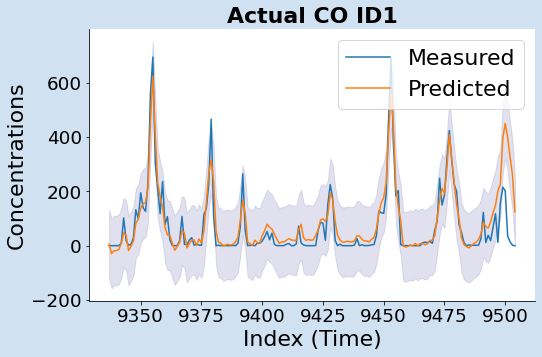}}
    \label{fig:cp_id1_co_7d}
    \hfil
    \subfloat[]{\includegraphics[width=.23\textwidth]{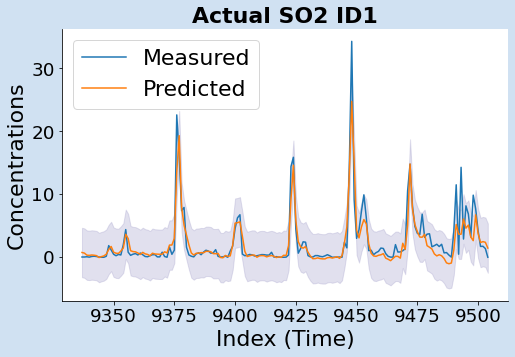}}
    \label{fig:cp_id1_so2_7d}
    \hfil
    \subfloat[]{\includegraphics[width=.23\textwidth]{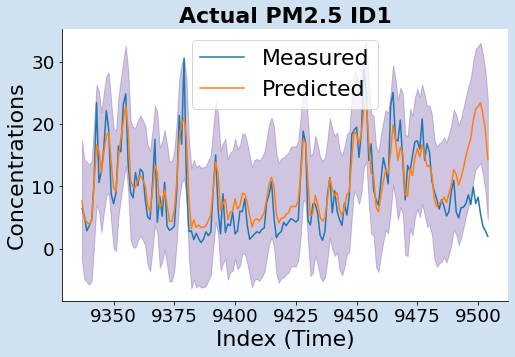}}
    \label{fig:cp_id1_pm2_7d}
    \hfil
    \subfloat[]{\includegraphics[width=.23\textwidth]{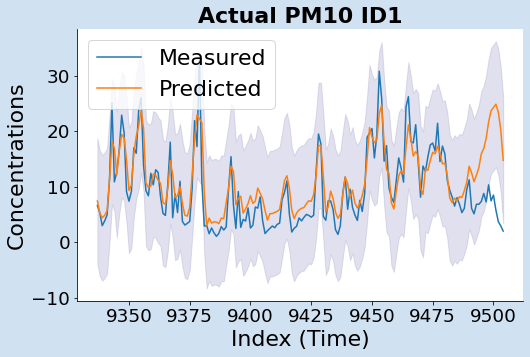}}
    \label{fig:cp_id1_pm10_7d}
    \caption{Actual vs Forecasts of the targeted pollutants for ID1 Sensor for 7 day forecast horizon (a) $NO_2$, (b) $O_3$, (c) $CO$, (d) $SO_2$, (e) $PM_{2.5}$, and (f) $PM_{10}$}
    \label{fig:cp 7d id1}
\end{figure}

\begin{table*}
\captionsetup{justification=centering}
    \caption{Performance of the proposed WaveCatBoost model and selected baseline forecasters for CPCB and ID1 sensors, evaluated based on the MASE metric for different forecast horizons.\label{Lumpmase}}
    \centering
    \resizebox{\textwidth}{!}{
    \begin{tabular}{|c|c|c|c|c|c|c|c|c|c|c|c|c|c|c|c|c|c|}
    \hline

    \multirow{2}{*}{Pollutant} & \multirow{2}{*}{Horizon} & \multicolumn{8}{c|}{CPCB Sensor} & \multicolumn{8}{c|}{ID1 Sensor}
        \\ \cline{3-10} \cline{11-18} & & RNN & LGBM & TCN & TFT & Transformer & XGB  & CatBoost & WaveCatBoost & RNN & LGBM & TCN & TFT & Transformer & XGB  & CatBoost & WaveCatBoost  \\ \hline
        \multirow{4}{4em}{$NO_2$} & 1d & 2.45 & 0.79 & 1.45 & 1.41 & 3.28 & 0.91 & 0.82 & \textbf{0.57} & 2.05 & 1.27 & 2.15 & 1.62 & 2.07 & 1.40 & 1.20 & \textbf{0.91} \\ 
        & 7d & 1.35 & \textbf{0.66} & 0.67 & 2.65 & 0.93 & \textbf{0.66} & 1.04 & 0.84 & 1.66 & 0.82 & 1.64 & 1.76 & 1.76 & 0.95 & 0.84 & \textbf{0.52}\\ 
        \textbf{} & 14d & 0.81 & \textbf{0.71} & 1.79 & 4.27 & 2.05 & 0.72 & 0.73 & 0.72 & 1.57 & 0.85 & 1.56 & 2.06 & 1.63 & 0.92 & 1.69 & \textbf{0.42} \\ 
        \textbf{} & 31d & 1.98 & 0.75 & 1.87 & 1.18 & 1.35 & 0.76 & 1.17 & \textbf{0.47} & 1.47 & 1.15 & 1.48 & 1.95 & 1.86 & 1.14 & 1.36 & \textbf{0.40} \\ \hline
        
        \multirow{4}{4em}{$O_3$} & 1d & 5.19 & 1.87 & 1.85 & 2.43 & 11.47 & 1.52 & 2.69 & \textbf{0.98} & 3.65 & 1.61 & 3.32 & 2.79 & 3.75 & 1.72 & 1.49 & \textbf{1.42}\\ 
        & 7d & 5.46 & 1.68 & 4.66 & 6.16 & 5.82 & 1.51 & 4.84 & \textbf{0.76}  & 3.57 & 1.64 & 2.55 & 2.84 & 3.18 & 1.96 & 1.79 & \textbf{0.72} \\ 
        & 14d & 6.24 & 1.23 & 4.38 & 14.68 & 9.54 & 1.27 & 1.77 & \textbf{1.11} & 3.85 & 1.80 & 3.67 & 3.74 & 3.12 & 2.06 & 2.32 & \textbf{0.64}\\ 
        & 31d & 3.56 & 2.46 & 4.26 & 2.71 & 3.56 & 2.44 & 5.37 & \textbf{0.25} & 3.59 & 2.72 & 3.63 & 3.56 & 3.61 & 2.87 & 2.66 & \textbf{1.11} \\ \hline
        
        \multirow{4}{4em}{$CO$} & 1d & 2.36 & 3.36 & 2.90 & 2.45 & \textbf{0.97} & 3.21 & 2.87 & 2.69 & 4.47 & 2.48 & 3.82 & 2.95 & 4.47 & 2.54 & 2.32 & \textbf{1.53}\\ 
        & 7d & 3.74 & 3.62 & 3.94 & 5.29 & 3.74 & 3.59 & 3.57 & \textbf{2.61} & 2.85 & 1.52 & 2.64 & 2.95 & 2.84 & 1.46 & 1.36 & \textbf{0.79}\\ 
        & 14d & 2.76 & 3.06 & 2.79 & 4.34 & 2.77 & 2.88 & 3.15 & \textbf{1.63} & 3.08 & 1.93 & 3.08 & 3.71 & 3.06 & 1.83 & 3.36 & \textbf{0.83}\\ 
        & 31d & 1.92 & 2.05 & 1.96 & 3.29 & 1.91 & 1.98 & 2.03 & \textbf{0.61} & 2.79 & 2.58 & 2.79 & 3.48 & 3.57 & 2.22 & 3.26 & \textbf{0.71} \\ \hline

        \multirow{4}{4em}{$SO_2$} & 1d & 34.35 & 11.46 & 28.11 & \textbf{7.76} & 40.15 & 10.62 & 9.35 & 23.93 & 1.81 & 0.45 & 0.85 & 0.49 & 1.79 & \textbf{0.43} & 0.44 & 1.31 \\ 
        & 7d & 16.83 & 15.43 & 14.62 & 20.72 & 11.59 & 14.39 & \textbf{5.84} & 7.30 & 2.36 & 0.34 & 2.23 & 0.61 & 2.21 & 0.33 & \textbf{0.30} & 0.61 \\ 
        & 14d & 10.70 & 7.35 & 9.37 & 24.75 & 9.57 & 8.14 & 11.31 & \textbf{6.17} & 2.15 & 0.22 & 2.82 & 1.66 & 3.18 & 0.26 & \textbf{0.24} & 0.58 \\ 
        & 31d & 9.06 & 11.22 & 10.34 & 11.44 & 10.79 & 9.13 & 9.81 & \textbf{5.33} & 2.35 & 0.41 & 2.95 & 0.81 & 1.88 & \textbf{0.26} & 0.31 & 1.31 \\ \hline
        
        \multirow{4}{4em}{$PM_{2.5}$} & 1d & 2.88 & 3.81 & \textbf{2.50} & 3.52 & 6.57 & 2.95 & 3.02 & 5.06 & 7.31 & 4.67 & 7.03 & 4.68 & 7.35 & 5.53 & 4.45 & \textbf{0.85}\\ 
        & 7d & 6.94 & 6.86 & 7.05 & 7.38 & 6.69 & 5.92 & 6.57 & \textbf{5.07} & 7.09 & 3.01 & 5.22 & 4.22 & 5.56 & 2.65 & 2.43 & \textbf{0.58}\\ 
        & 14d & 4.40 & 6.98 & 4.40 & 7.07 & 4.41 & 6.20 & 6.30 & \textbf{2.34} & 8.20 & 4.43 & 8.71 & 7.10 & 7.39 & 3.67 & 5.58 & \textbf{0.69}\\ 
        & 31d & 2.73 & 2.85 & 2.83 & 4.41 & 2.73 & 2.83 & 2.83 & \textbf{1.57} & 7.21 & 5.26 & 7.01 & 5.97 & 5.05 & 4.49 & 5.29 & \textbf{0.42} \\ \hline
        
        \multirow{4}{4em}{$PM_{10}$} & 1d & \textbf{1.01} & 6.69 & 1.79 & 6.37 & 4.16 & 3.81 & 5.77 & 3.12 & 7.31 & 4.34 & 7.06 & 4.24 & 7.35 & 5.59 & 4.22 & \textbf{0.86}\\ 
        & 7d & 5.20 & 5.30 & 5.32 & 6.19 & 5.11 & 4.99 & 5.24 & \textbf{3.70} & 7.37 & 2.83 & 5.42 & 4.34 & 5.81 & 2.46 & 2.27 & \textbf{0.57}\\ 
        & 14d & 3.53 & 4.59 & 3.55 & 5.90 & 3.54 & 4.36 & 4.89 & \textbf{1.48} & 8.48 & 4.29 & 9.08 & 7.08 & 7.65 & 3.29 & 5.44 & \textbf{0.67} \\ 
        & 31d & 2.39 & 2.40 & 2.51 & 3.64 & 2.37 & 2.43 & 2.40 & \textbf{0.98} & 7.57 & 5.39 & 7.35 & 6.09 & 5.36 & 4.50 & 5.47 & \textbf{0.41} \\ \hline
    \end{tabular}}
\end{table*}

Table \ref{Lumpmase} displays the performance of selected forecasting architectures, incorporating only the best those that perform in at least one scenario. For the CPCB sensor, the MASE metric values in the table indicate that the WaveCatBoost model outperforms all the baseline architectures for long-range forecasting periods of 14 days and 31 days for most air pollutants. This improved performance is attributed to the hybrid training approach, where MODWT aids in extracting signals from noise, and CatBoost accurately extrapolates these transformed signals. In the 7-day forecasting scenario, WaveCatBoost performs comparably to other boosting methods like LGBM, XGB, and CatBoost across various pollutants. Furthermore, for short-term forecasts with a 1-day horizon, WaveCatBoost demonstrates competitive performance with transformer-based architectures. Additionally, the MASE metric for ID1 sensors in Table \ref{Lumpmase} indicates that our proposed model outperforms all baseline models across different horizons for all air pollutants except for SO2, where XGB and CatBoost models offer competitive results.\par 
Furthermore, we study the statistical significance of the improvement in model performance using multiple comparisons with the best (MCB) test procedure \cite{koning2005m3}. This distribution-free test ranks each model based on its MASE metric across different air pollutants and computes their average rank. The model with the lowest MASE metric is identified as the ``best'' method, determined by achieving the minimum rank. The MCB plots depicted in Fig. \ref{fig:mcb fh} illustrate that WaveCatBoost models exhibit superior performance for both the ID1 (Fig. \ref{fig:mcb_id1}) and CPCB (Fig. \ref{fig:mcb_cpcb}) sensors. Additionally, the boundary of the critical distance (blue line) of the best-performing model serves as the reference value (shaded region) for the test. Given that most of the baseline forecasters have non-overlapping critical distance (blue line) with the reference value of the test, we can conclude that their performance is significantly inferior to that of the proposed WaveCatBoost architecture. The overall experimental analysis underscores the effectiveness of our proposed model in outperforming a diverse range of benchmark models, emphasizing its potential as a reliable tool for real-time air quality forecasting. To quantify the uncertainty in the WaveCatBoost forecasts, we utilize the conformal prediction approach as in Eq. \ref{CP_WCB_Eq}. Figure \ref{fig:cp 1d id1} and \ref{fig:cp 7d id1} 
showcase the point forecast generated by the WaveCatBoost model, ground truth, and the conformal prediction intervals at $\alpha = 0.05$ for the 1-day and 7-day horizon of ID1 sensors, respectively (as it gives a probabilistic band, therefore sometimes the band consists of negative values which for this case should be considered as zero). From the plot, it is evident that the proposed framework is generalizable and robust, offering valuable insights for environmental monitoring and public health interventions. 

\section{Conslusion}
In this letter, we designed a WaveCatBoost model to forecast real-time air pollutant concentration levels. These air pollutants, including gaseous and particulate matter, lead to several carcinogenic and non-carcinogenic health hazards. Our architecture integrates wavelet decomposition with the CatBoost approach to effectively capture the non-stationary and long-term dependencies inherent in pollutant time series data. Experimental evaluation using two real-world datasets from Meghalaya, India, demonstrates that our proposal outperforms baseline forecasting methods. Furthermore, we assess the robustness of the model through statistical significance tests and provide probabilistic bands for our forecasts based on a conformal prediction approach. These findings are crucial for advancing the field of air quality forecasts and guiding future research endeavors to enhance predictive models for environmental sustainability. 
As a future direction, further analysis of the WaveCatBoost architecture in forecasting air pollutant levels by considering spatial dependencies emerges as a promising avenue in air quality research.


\bibliographystyle{IEEEtran}
\bibliography{ref}

\end{document}